\documentclass[11pt]{article}

\usepackage[preprint]{acl}

\usepackage{times}
\usepackage{latexsym}

\usepackage[T1]{fontenc}

\usepackage[utf8]{inputenc}

\usepackage{microtype}

\usepackage{inconsolata}

\usepackage{graphicx}

\title{DOSA: A Tree-Guided, Self-Regressive Framework for Long Document Structure Analysis}

\author{
Bohou Li \and Benjamin Sowell \and Mehul Shah \and Mark Lindblad \and Henry Lindeman \\
Glean Technologies, Inc.
}

\begin{document}
\maketitle
\begin{abstract}
In visually-rich documents, information is encoded not only in individual page objects such as tables, headers, and text blocks, but also in the structural relations among them, making document structure analysis fundamental to information retrieval and document understanding. However, accurately inferring such relations remains challenging in multi-page documents with long-range dependencies and heterogeneous layouts. To address this, we propose a tree-guided and self-regressive framework, termed \textbf{DOcument Structure Analyzer (DOSA)}, for inferring relations among page objects and reconstructing document-level semantic trees. DOSA processes documents chunk-by-chunk, fusing visual, textual, and layout features for each page object and predicting hierarchical and ordering relations. The predicted relations are used to incrementally construct a semantic tree, which is then leveraged as structural context to guide inference on subsequent chunks. Experimental results on five benchmarks demonstrate the effectiveness of DOSA, with improvements of up to 4 F1 points and 19 TEDS points on DocHieNet, the most challenging multi-page hierarchy benchmark.
\end{abstract}

\section{Introduction}
\begin{figure*}[t]
\centering
\includegraphics[width=\textwidth]{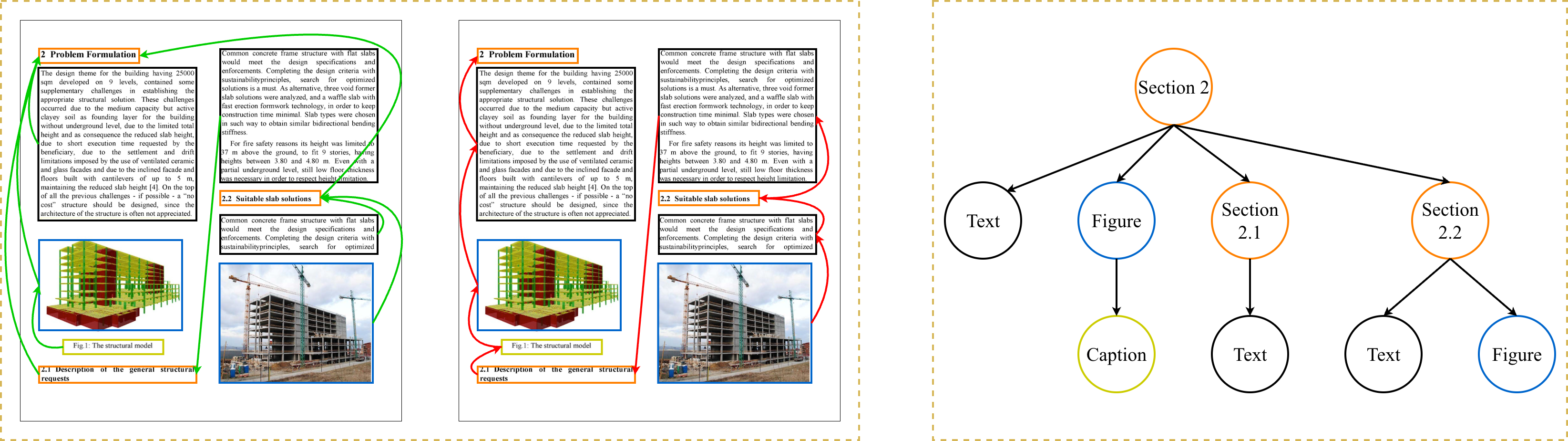}
\caption{Example document structural relations and the corresponding semantic tree. Green edges denote parent relations and red edges denote predecessor relations. In the resulting semantic tree, Section 2 is the parent of a paragraph, a figure, and subsections 2.1 and 2.2. The figure further contains a caption node; subsection 2.1 contains a text paragraph; and subsection 2.2 contains a text node and another figure node. Siblings under the same parent are organized left to right in document reading order.}

\label{structure}
\end{figure*}

Today, large volumes of data are generated and stored in visually-rich document formats such as PDFs and slide decks. Unlike raw text content, these documents consist of page objects that exhibit rich logical, spatial, and semantic relations. As shown in Figure~\ref{structure}, the document structure analysis (DSA) task aims to uncover these relations and construct a semantic tree, a rooted tree over page objects in which edges represent parent relations and sibling nodes are ordered according to reading order. Such structural understanding is fundamental to traditional information retrieval, and also plays a critical role in modern applications, including accurate document chunking for retrieval-augmented generation (RAG) systems, structured property extraction, and coherent, hierarchy-aware document summarization through the establishment of clear semantic boundaries.

Early work on document structure analysis \citep{rausch2023dsg, chen2025graphdoc} primarily focused on single-page documents and relied solely on visual features. Subsequent approaches \citep{ma2023hrdoc, xing2024dochienet} extended this line of research to multi-page documents, employing transformer-based architectures to model interactions among page objects using multimodal features. These methods achieved better performance on multi-page benchmarks. However, transformers introduce significant computational overhead and can suffer from accuracy degradation due to attention dilution in long documents with many page objects. Cross-page modeling thus remains non-trivial, particularly when long-range dependencies and diverse page layouts must be reconciled.

To address these challenges, we propose \textbf{DOcument Structure Analyzer (DOSA)}, a unified framework that reformulates long-document structure analysis as an incremental, self-regressive semantic tree construction problem. Unlike prior approaches that primarily treat the semantic tree as the final prediction target, DOSA additionally leverages the partially constructed semantic tree as structural context for subsequent prediction. Instead of processing an entire document in a single pass, DOSA operates in a chunk-by-chunk manner. Each chunk is processed by a multimodal model designed for document structure parsing. For each page object, the model extracts and fuses visual, semantic, categorical, and sizing features. Candidate objects are organized into structured sequences and encoded by a transformer equipped with global positional encoding that integrates page indices and bounding box coordinates. The resulting representations are used to predict hierarchical and ordering relations within the chunk. These predicted structural relations are then used to update the semantic tree, which dynamically guides subsequent inference by identifying structurally relevant page objects as context for the next chunk's prediction. This iterative process continues until all chunks have been processed. By conditioning each step on structurally relevant candidates, DOSA mitigates the sequence-length and attention-dilution limitations of transformers while preserving global structural coherence.

In summary, we make the following key contributions:
\begin{itemize}
\item We propose a unified framework that reformulates document structure analysis as an incremental, self-regressive semantic tree construction process, where partially constructed semantic trees are leveraged as structural context for subsequent prediction, enabling scalable and coherent parsing of long documents.
\item We design an efficient model tailored for document structure analysis. The model integrates visual, semantic, categorical, and sizing features to predict hierarchical and ordering relations, achieving strong performance without relying on large-scale document-specific foundation model pretraining.
\item We conduct extensive experiments on five document structure parsing datasets, demonstrating that DOSA outperforms state-of-the-art methods by a large margin.
\item We provide analyses and ablations on DocHieNet, showing that DOSA surpasses large language model baselines (Gemini-2.5-Pro, GPT-5.2) and validating the contributions of tree-guided context selection and multimodal feature fusion.
\end{itemize}

\begin{figure*}[t]
\centering
\includegraphics[width=\textwidth]{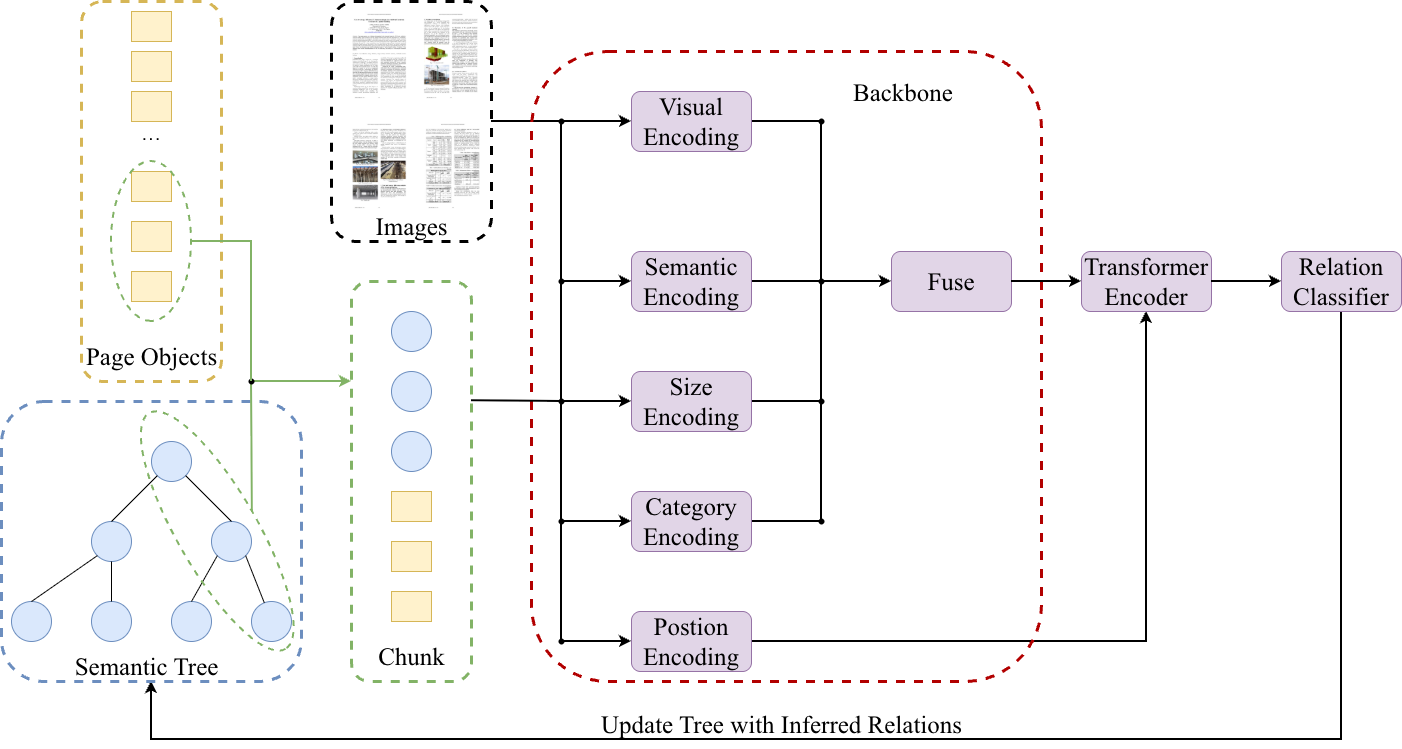}
\caption{Overview of DOSA. DOSA selects context page objects from the rightmost branches of the current semantic tree and merges them with an unresolved chunk. The resulting sequence is processed by the multimodal backbone, a transformer encoder, and relation prediction heads to infer hierarchical and ordering relations. The inferred relations are then used to incrementally update the semantic tree. This process continues until the entire document has been processed.}
\label{architecture}
\end{figure*}

\section{Related Work}

\subsection{Document Structure Analysis}
Early datasets for document hierarchy analysis, such as arXivDocs \citep{rausch2021docparser} and E-Periodica \citep{rausch2023dsg}, focused on single-page documents with relatively limited layout and linguistic diversity. GraphDoc \citep{chen2025graphdoc} extended this line of work by augmenting the more diverse dataset DocLayNet \citep{pfitzmann2022doclaynet} with explicit relational annotations. However, single-page datasets cannot capture structural dependencies that span across pages, which are crucial for comprehensive document hierarchy analysis. To address this, HRDoc \citep{ma2023hrdoc} and DocHieNet \citep{xing2024dochienet} shifted the focus to multi-page settings, where long-range dependencies become essential.

Modeling approaches evolved alongside these datasets. Early methods such as DocParser \citep{rausch2021docparser}, which relied on heuristic rules, had limited generalizability, while DSG \citep{rausch2023dsg} adopted a bidirectional LSTM but incorporated only visual features. DRGG \citep{chen2025graphdoc} introduced semantic graph modeling and citation relations, yet remained restricted to single-page documents and primarily visual representations. For multi-page settings, \citet{ma2023hrdoc} proposed DSPS, which combined transformer-based page encoding with a structure-aware GRU \citep{cho2014properties} decoder for cross-page relation detection. Building on this framework, \citet{wang2024detect} introduced DOC, which decomposed document structure analysis into within-page ordering and section-header hierarchy stages, leading to improved performance on HRDoc. More recently, DHFormer \citep{xing2024dochienet} integrated fine-grained textual semantics with coarse layout patterns for multi-page hierarchy modeling.

\subsection{Document Transformers}
The transformer architecture \citep{vaswani2017attention} enables effective modeling of long-range dependencies and efficient parallel training, making it well suited for capturing relations among page objects in document structure analysis. However, the quadratic complexity of self-attention leads to computational and memory bottlenecks as sequence lengths grow, and attention dilution \citep{qin2022devil,han2024rag} can reduce focus on relevant elements. DSPS employed a GRU for cross-page information, but per-page transformer encoding restricted object-level dependencies across pages. DOC mitigated sequence-length issues by modeling hierarchical relations only between section headers; however, this relied heavily on accurate header classification and ignored potential hierarchical relations among non-heading objects (e.g., lists or plain text). MultiDocFusion \citep{shin2025multidocfusion} took a similar approach using instruction-tuned LLMs. DHFormer preserved dense self-attention within pages while refining page-object representations globally in the decoder, using shifted sparse attention (SSA) \citep{chen2023longlora} to handle longer sequences. Although SSA alleviates sequence-length pressure, sparse attention inherently limits attended positions and may miss fine-grained object-level relations.

Beyond document modeling, substantial prior work has focused on alleviating the sequence length limitation of transformers by modifying attention mechanisms. Common strategies include sparse or local-global attention \citep{beltagy2020longformer} and hierarchical or chunk-based processing, where long documents are segmented, encoded independently, and later aggregated \citep{pappagari2019hierarchical}. These methods avoid full pairwise attention by preselecting a subset of tokens or regions deemed relevant. However, such approaches often lack task-specific knowledge for locating the most relevant context. We argue that effective long-document modeling requires a task-aware mechanism to precisely identify candidate page objects for attention, instead of depending on the model's implicit attention patterns or fixed configurations.

\begin{figure*}[t]
  \centering
  \includegraphics[width=\textwidth]{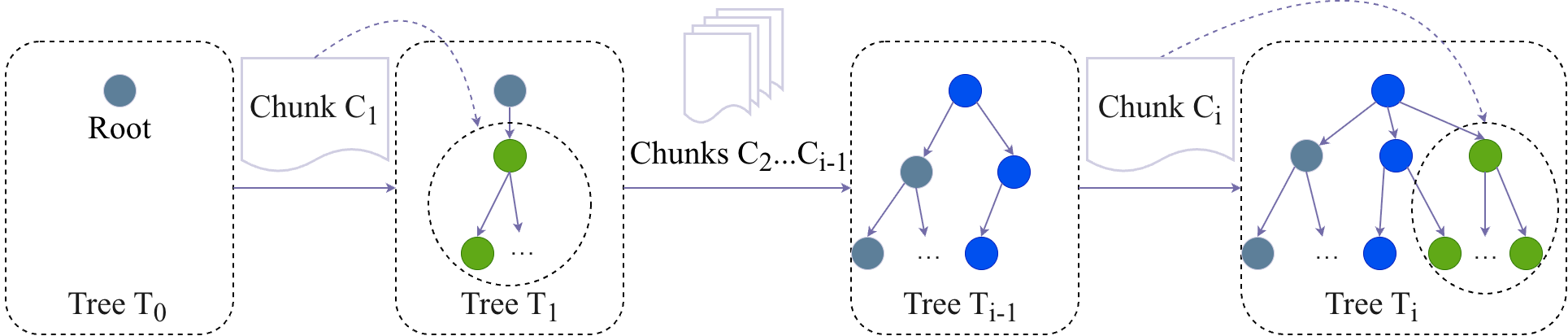}
  \caption{Incremental construction of the semantic tree from inferred relations. The green nodes in $T_1$ represent page objects whose relations are resolved from chunk $C_1$. The blue nodes in $T_{i-1}$ denote the rightmost branch of the partially constructed tree. As illustrated by the transition from $T_{i-1}$ to $T_{i}$, newly processed page objects can only relate to nodes on the rightmost branch of $T_{i-1}$ and are therefore appended to the right of this branch.}
  \label{semantic}
\end{figure*}

\section{Problem Definition}
We treat document structure analysis as a downstream task for document object detection and text extraction. The input to the framework is a multi-page document along with N extracted page objects $\{O_1, O_2, \cdots, O_N\}$. Each page object is characterized by its category, bounding box location, and the textual content within the bounding box. The framework models two types of structural relations: hierarchical relations, captured by parent–child links between page objects, and ordering relations, captured by identifying the immediately preceding page object in the document reading order.

\section{Self-Regressive Semantic Tree}
As discussed, transformer-based models are effective at capturing long-range dependencies but face practical limitations when processing long documents with many page objects. Rather than modifying the transformer architecture to directly accommodate longer input sequences, we propose a self-regressive semantic-tree-based approach for scalable long-document processing. Our method incrementally constructs a semantic tree and leverages the partial semantic tree as a prior to restrict the set of page objects involved in attention, thereby reducing computational cost while preserving relevant contextual information.

\subsection{Semantic Tree Construction}
Given a long document, we first partition it into $M$ chunks. Each chunk contains one or more pages with up to $N$ page objects, where $N$ corresponds to the maximum input sequence length of the transformer encoder. If a chunk contains fewer than $N$ page objects, we pad it with empty tokens. Starting from the first chunk, we perform page-object relation detection sequentially while incrementally constructing a semantic document tree. As illustrated in Figure~\ref{semantic}, the semantic tree $T_i$ is obtained by updating the previous tree $T_{i-1}$ using the relation detection results from chunk $C_i$. Inductively, $T_i$ represents a partial semantic tree constructed from chunks $C_1, C_2, \ldots, C_i$. After processing all $M$ chunks, we obtain a complete semantic tree corresponding to the entire document.

Ideally, each page object has a single parent and a well-defined ordering with its predecessor, yielding a clean tree structure. In practice, relation classification errors may introduce structural inconsistencies, such as cycles where two objects are each predicted to be the parent of the other. To address this, we apply a simple decycling procedure that removes the lowest-confidence edge. Conflicts may also arise between parent--child and ordering predictions; for example, two children $o_1$ and $o_2$ under the same parent may lack a transitive ordering relation. In such cases, we fall back to spatial ordering based on bounding box positions.

\subsection{Tree-Guided Context Selection}
A key limitation of the chunk-based processing described above is that page objects in earlier chunks may still exhibit structural relations with objects in subsequent chunks. Naively processing chunks in isolation would therefore ignore such cross-chunk dependencies.

Documents exhibit an intrinsic ordering induced by page sequence and structural relations among page objects. The incrementally constructed semantic tree preserves the document's structural ordering. In particular, a depth-first traversal of the tree corresponds to the document's reading order, where objects on earlier pages precede those on later pages. This property enables efficient context selection when processing a new chunk. Suppose that at step $i-1$ we have a partially constructed semantic tree $T_{i-1}$ and are about to process a new chunk $C_i$. If a page object $P_{child} \in C_i$ has a parent $P_{parent}$ already present in $T_{i-1}$, then $P_{parent}$ must lie on the rightmost branch of $T_{i-1}$. The formal proof is provided in Appendix \ref{sec:appendix_rightmost}. Similarly, if $P_{child}$ has a predecessor in $T_{i-1}$, that predecessor must be the last object in $T_{i-1}$. This invariant implies that the rightmost branch of $T_{i-1}$ is sufficient as context for processing chunk $C_i$.

In practice, however, the semantic tree may contain errors due to imperfect relation classification, and such errors can accumulate across chunks (see Appendix~\ref{sec:appendix_error_accumulation}). To improve robustness, we introduce a rightmost-branches-based soft window mechanism. Specifically, context objects are first retrieved from the rightmost branch of the partial semantic tree $T_{i-1}$. If the predefined window size is not yet reached, objects are then added from the second rightmost branch, followed by the third, and so on, until the window is filled. The selected context objects are merged with the current chunk and provided to the model for inference. This design mitigates early prediction errors by preserving multiple structurally plausible candidates in context, reducing sensitivity to local misclassifications.

In summary, by leveraging the partially constructed semantic tree as a structural prior, we restrict attention to a compact yet informative set of context objects when processing each new chunk. This design enables the transformer to model long-range cross-page dependencies through attention while avoiding the prohibitive cost of full-document encoding.

\section{In-Chunk Model Architecture}
The architecture of the model is conceptually simple and consists of three components: (1) a feature extraction backbone that produces multimodal representations for each page object, (2) a transformer encoder that refines these representations by modeling contextual interactions among page objects, and (3) a multilayer perceptron (MLP) that predicts the final structural relations.

\subsection{Backbone}
The DOSA backbone extracts and encodes multimodal features for each page object, including visual appearance, semantic content, category information, and sizing cues.

\paragraph{Visual Embedding.} To process an input image of size $3 \times H \times W$ (with 3 color channels), we employ a standard ResNet-50 FPN \citep{lin2017feature} to construct feature pyramids. For each page object, we use the MultiScaleROIAlign algorithm to extract feature maps of size $256 \times 7 \times 7$ (with 256 channels) from FPN layers $\{P_2, P_3, P_4, P_5, P_6\}$. Each feature map is flattened into a one-dimensional sequence and passed through a linear projection layer to produce a 512-dimensional embedding, which helps prevent the visual modality from dominating the joint representation during fusion. The final visual embedding is computed as:
\begin{equation}
    E_{v}=\mathrm{FC}(\mathrm{Flatten}(\mathrm{ROI}(F, B)))
\end{equation}
where $F$ denotes the feature map extracted from the FPN, $B$ represents the bounding box tensor, and $E_{v}$ is the resulting visual embedding.

\paragraph{Semantic Embedding.} We employ a straightforward strategy to represent the semantics of each page object. If the page object contains text, we use it as the content; otherwise, we fall back on the object's category—for example, a figure is represented as ``A figure.'' This ensures that every page object has a consistent semantic representation. The resulting content is then encoded using the pre-trained language model BERT \citep{devlin2019bert}.

\paragraph{Category Encoding.} The category of a page object provides useful cues for predicting relations between objects; for example, a section header followed by a text paragraph often implies a parent--child relation. We represent the category as an integer and employ a simple fully connected layer to map this 1-dimensional input to a 128-dimensional embedding.

\paragraph{Sizing Embedding.} Page object size can be informative for hierarchy prediction, especially when section numbers are absent from different header levels. However, ROIAlign \citep{he2017mask} does not explicitly preserve the absolute dimensions of objects within the extracted feature maps. When feature maps are extracted for objects of different sizes, ROIAlign resizes them into a fixed grid (e.g., $7 \times 7$), which captures spatial alignment but discards absolute sizing information. To retain this information, we extract the width, height, and area of each page object and normalize them by the corresponding page width, height, and area to produce a 3-dimensional sizing vector. This vector is then projected into a 128-dimensional embedding using a linear layer, similar to the category encoding.

\paragraph{Fusion.} To obtain the final page object representation, we concatenate the visual, semantic, category, and sizing embeddings and pass them through a two-layer MLP. This projects the concatenated embedding into a 1024-dimensional space and adjusts the relative importance of each feature. Formally, the fused embedding is:
\begin{equation}
    E_{f} = \mathrm{MLP}(E_v \oplus E_s \oplus E_c \oplus E_z),
\end{equation}
where $E_v$, $E_s$, $E_c$, and $E_z$ denote the visual, semantic, category, and sizing embeddings, respectively.

\subsection{Transformer Encoder}
We treat each page object embedding as a token, and a sequence of $N$ page objects forms the input to the transformer. A 6-layer transformer encoder is employed to enhance the fused page object representations by capturing contextual information from neighboring objects. Each encoder layer follows the standard architecture, consisting of a multi-head self-attention module and a feed-forward network (FFN). Since transformers are permutation-invariant, we incorporate position encodings to provide order and spatial information for each page object.

\paragraph{Position Encoding.} For single-page relation detection, a page object's bounding box is typically sufficient to capture positional information. However, this is inadequate when modeling relations across multiple pages. To address this, we construct a global position vector by combining the bounding box coordinates with the page number. Document lengths can vary significantly, from a single page to thousands of pages. In DOSA, long documents are split into chunks during both training and inference. To keep page numbers within a small, stable range and avoid extreme outliers, we use the relative page number within each chunk instead of the absolute page number in the full document. Specifically, the position vector for each page object is defined as $[page\_no, x_1, y_1, x_2, y_2]$, where $(x_1, y_1)$ and $(x_2, y_2)$ are the coordinates of the upper-left and lower-right corners of the bounding box, and $page\_no$ is the relative page index within the chunk. This vector is normalized by the page width, page height, and total page count of the chunk. Finally, a fully connected layer with 5-dimensional input and 1024-dimensional output maps the vector into the position encoding used by the transformer.

\subsection{Relation Classifier}
Similar to DOC \citep{wang2024detect}, DOSA employs a multi-class classifier to predict the parent and the immediate predecessor for each page object. With the exception of the root node and the first object in a document, every page object has exactly one parent and one predecessor. The root node has no parent, and the first object has no predecessor; to maintain representational consistency, we assign both relations to the object itself in these cases.

Given a chunk $C$ containing $N$ page objects ${o_1, o_2, \ldots, o_N}$, we construct $N$ candidate relation embeddings for each target object $o_i$ by concatenating its representation with that of every object in the chunk, including itself, as $\{o_1 \oplus o_i, o_2 \oplus o_i, \ldots, o_N \oplus o_i\}$. These concatenated embeddings are passed through a two-layer multilayer perceptron (MLP) to produce logits ${z_{1i}, z_{2i}, \ldots, z_{Ni}}$, where $z_{ji}$ represents the confidence score of object $o_j$ being related to $o_i$. A softmax function is then applied to obtain a probability distribution over all candidates, and the object with the highest probability is selected.
\begin{equation}
    \arg\max_{j \in {1, \dots, N}} \left( \frac{e^{z_{ji}}}{\sum_{k=1}^{N} e^{z_{ki}}} \right)
\end{equation}
To train the relation classifier, we use the focal loss \citep{lin2017focal}, which mitigates potential class imbalance by emphasizing ambiguous cases with competing candidates.

\section{Experiments}

\begin{table*}[t]
\centering
\setlength{\tabcolsep}{6pt}
\begin{tabular}{lcccccccccc}
\hline
Dataset &
\multicolumn{2}{c}{\textbf{arXivDocs}} &
\multicolumn{2}{c}{\textbf{HRDS}} &
\multicolumn{2}{c}{\textbf{HRDH}} &
\multicolumn{2}{c}{\textbf{E-Periodica}} &
\multicolumn{2}{c}{\textbf{DocHieNet}} \\
\cline{2-3}
\cline{4-5}
\cline{6-7}
\cline{8-9}
\cline{10-11}
 & F1 & TEDS & F1 & TEDS & F1 & TEDS & F1 & TEDS & F1 & TEDS \\
\hline
DocParser & 58.14 & 29.11 & 56.84 & 28.71 & 47.36 & 22.39 & 35.20 & 18.67 & 23.31 & 6.81 \\
DSPS      & --    & --    & --    & 81.74 & --    & 69.71 & --    & --    & --    & --   \\
DOC       & --    & --    & --    & 95.10 & --    & 85.48 & --    & --    & --    & --   \\
DSG       & 81.17 & 72.47 & 84.78 & 83.24 & 74.04 & 64.33 & 67.17 & 60.14 & 53.51 & 33.90 \\
DHFormer  & \textbf{99.70} & 97.42 & \textbf{99.57} & 97.98 & 96.69 & 92.63 & 95.76 & 93.09 & 77.82 & 57.64 \\
\hline
DOSA  & 98.94 & \textbf{97.68} & 99.41 & \textbf{99.34} & \textbf{96.75} & \textbf{96.52} & \textbf{98.03} & \textbf{94.77} & \textbf{81.80} & \textbf{76.81} \\
\hline
\end{tabular}
\caption{Performance comparison across different datasets measured by F1 and TEDS.}
\label{tab:performance}
\end{table*}

\subsection{Implementation Details}
The DOSA framework is compact and efficient, with approximately 33M trainable parameters, enabling training and inference on modest hardware. All experiments are conducted on a single workstation equipped with one NVIDIA L4 GPU (24\,GB memory). The visual backbone is a ResNet-50 \citep{he2016deep} pretrained on ImageNet, and the semantic backbone is sentence-transformers/distiluse-base-multilingual-cased-v2; parameters of both backbones are frozen during training. Transformer layers are initialized using Xavier initialization \citep{glorot2010understanding}. We optimize the model using AdamW \citep{loshchilov2018decoupled} with a mini-batch size of 2, an initial learning rate of $5 \times 10^{-5}$, and weight decay of $10^{-4}$. By default, training is performed for 100 epochs, with the learning rate decayed by a factor of 10 after 90 epochs. AdamW hyperparameters are set to $\beta_1=0.9$, $\beta_2=0.999$, and $\epsilon=10^{-8}$. The context window size is set to 8. We use the same training configuration across all datasets, with minor adjustments to the number of epochs and window size based on dataset characteristics. Additional implementation details are provided in Appendix~\ref{sec:appendix_training}.

\subsection{Evaluation Metrics}
Following prior work \citep{rausch2023dsg,xing2024dochienet}, we use F1-score to evaluate page object relation prediction and Tree-Edit-Distance-based Similarity (TEDS) \citep{zhong2020image,zhong2019publaynet} to measure predicted and ground-truth semantic tree similarity.

\subsection{Datasets}
We conduct experiments on five widely-used document structure datasets: two single-page datasets (arXivDocs and E-Periodica) and three multi-page datasets (HRDS, HRDH, and DocHieNet), where HRDS and HRDH are the simple and hard splits of HRDoc. To ensure compatibility with DOSA, we apply a preprocessing pipeline to standardize the inputs; details are provided in Appendix~\ref{sec:appendix_dataset}. DocHieNet is particularly challenging due to its diverse document domains and complex multi-page layouts, making it well-suited for revealing performance differences across model variants. It is therefore used for ablation and analysis experiments.

\subsection{Comparison with Document Models}
Table~\ref{tab:performance} presents the evaluation results of DOSA and existing methods. For a fair comparison with other models that assume access to ground-truth reading order, we provide DOSA with the ground-truth ordering for training and inference. DHFormer represents the current state of the art for document hierarchy detection. On the DocHieNet benchmark, DOSA substantially outperforms DHFormer, achieving gains of approximately 4 F1 points and nearly 20 TEDS points, demonstrating its effectiveness in modeling high-level document structure. On arXivDocs and HRDS, DHFormer achieves slightly higher F1 scores, while DOSA attains higher TEDS scores on both datasets, indicating more accurate overall semantic tree reconstruction.

\subsection{Comparison with Large Language Models}

\begin{table}[h]
\renewcommand{\arraystretch}{1.2}
\centering
\begin{tabular}{lcc}
\hline
\textbf{Model} & \textbf{F1} & \textbf{TEDS} \\
\hline
Gemini-2.5-Pro   & 65.58 & 47.18 \\
GPT-5.2 & 69.53 & 57.82 \\
DOSA    & \textbf{81.80} & \textbf{76.81} \\
\hline
\end{tabular}
\caption{Comparison with large language models.}
\label{tab:llm}
\end{table}

Recent LLMs demonstrate strong long-context reasoning capabilities. However, structured hierarchical document parsing requires explicit modeling of parent-child relations, and implicit sequential reasoning does not necessarily yield accurate tree reconstruction. We evaluate Gemini-2.5-Pro and GPT-5.2 on DocHieNet using inputs aligned with our standard benchmark setting, including both document images and structured text annotations, to ensure a fair comparison. The models are prompted to generate hierarchical structures in a unified output format, which are evaluated using F1 and TEDS. Further setup details are provided in Appendix~\ref{sec:appendix_llm}. As shown in Table~\ref{tab:llm}, DOSA consistently outperforms both Gemini-2.5-Pro and GPT-5.2 on both metrics. Although these models exhibit strong generative reasoning ability, the results suggest that explicit structural modeling is more effective for accurate hierarchical document reconstruction than purely generative approaches.

\subsection{Analysis of Context Construction}

\begin{table}[h]
\centering
\begin{tabular}{lcc}
\hline
\textbf{Strategy} & \textbf{F1} & \textbf{TEDS} \\
\hline
No Context & 81.61 & 56.30 \\
Sequential Window & 83.13 & 59.59 \\
Rightmost Branch & 83.89 & 71.96 \\
DOSA & \textbf{83.97} & \textbf{73.68} \\
\hline
\end{tabular}
\caption{Comparison of context selection strategies.}
\label{tab:context_strategy}
\end{table}

We conduct ablation studies to evaluate the effectiveness of tree-guided context construction. Table~\ref{tab:context_strategy} compares different strategies for retrieving context objects from previously processed chunks on DocHieNet benchmark documents containing more than one chunk.

Without any context, performance drops substantially—particularly in TEDS—highlighting the importance of modeling cross-chunk dependencies for coherent document structures. The sequential window strategy retrieves the last 16 objects in reading order from the partial semantic tree, yielding only marginal improvements. This indicates that simple positional proximity is insufficient for capturing hierarchical relations. Selecting objects from the single rightmost branch of the semantic tree significantly improves both F1 and TEDS, demonstrating that structurally guided retrieval is considerably more effective than flat sequential context. However, its F1 and TEDS scores are still lower than those of DOSA, suggesting that while the rightmost branch often provides accurate local predictions, it is more vulnerable to early structural errors, which can propagate and compromise overall tree consistency. DOSA's rightmost-branches-based soft window mechanism further improves F1 and TEDS by expanding the context branch-by-branch from the rightmost branch toward earlier branches until the predefined window size is reached. By preserving multiple structurally plausible candidates, this strategy enhances robustness to error accumulation and produces more globally consistent document structures. Additional analysis of window size sensitivity is provided in Appendix~\ref{sec:appendix_window_size}.

\subsection{Analysis of Feature Embedding}
We further analyze the impact of category embeddings, sizing embeddings, and reading-order supervision on model performance. The baseline uses category and sizing embeddings with ground-truth reading-order priors. Results are shown in Table~\ref{tab:feature_ablation}.

\paragraph{Category Embedding.}
To evaluate the contribution of category information, we remove the category embedding from the feature fusion stage. This leads to a substantial performance drop, with F1 decreasing by 11.67 points and TEDS by 14.54 points. The results indicate that accurate page-object categorization is critical for reliable relation prediction.

\paragraph{Sizing Embedding.}
We next examine the role of sizing cues by removing the sizing embedding from the fused representation. This leads to a 0.36-point decrease in F1 and a 0.42-point decrease in TEDS, indicating that size information provides a modest but consistent benefit for hierarchy detection. To further analyze the impact of sizing cues, we include a qualitative analysis in Appendix~\ref{sec:appendix_sizing_analysis}.

\paragraph{Reading Order.}
DOSA is capable of predicting ordering relations jointly with hierarchical relations using two prediction heads. In this setting, the semantic tree is constructed using inferred reading-order relations instead of ground-truth priors. This results in a decrease of 1.07 F1 points and 11.69 TEDS points compared to the baseline, highlighting the importance of accurate reading-order information. Despite this degradation, DOSA with inferred ordering still outperforms the current state-of-the-art method DHFormer, which depends on ground-truth reading order, demonstrating the robustness of our proposed framework even without access to this oracle signal.

\begin{table}[h]
\renewcommand{\arraystretch}{1.2}  
\centering
\begin{tabular}{lcc}
\hline
\textbf{Model} & \textbf{F1} & \textbf{TEDS} \\
\hline
DOSA (full model)             & \textbf{81.80} & \textbf{76.81} \\
Without category embedding    & 70.13 & 62.27 \\
Without sizing embedding      & 81.44 & 76.39 \\
With inferred order           & 79.73 & 65.12 \\
\hline
\end{tabular}
\caption{Feature embeddings ablation on DocHieNet.}
\label{tab:feature_ablation}
\end{table}

\section{Conclusion}
In this paper, we present DOSA, a novel framework for page-object relation detection and document structure analysis. By combining a self-regressive, tree-based context construction algorithm with a lightweight multimodal model tailored for in-chunk relation prediction, DOSA effectively models long-range structural dependencies across pages. Experimental results highlight the effectiveness of our approach in capturing multi-page, hierarchical document structures, and suggest its applicability to other long-document understanding tasks.

\section*{Limitations}
Despite its strong performance, DOSA has several limitations that suggest directions for future work. First, the current framework does not model citation or reference relations. While such relations are less common than hierarchical or ordering relations, supporting them would require explicitly detecting reference objects and incorporating them into the context construction process. Extending DOSA to handle graph-based relations beyond tree structures is a promising direction for future research. Second, although the soft window mechanism alleviates error propagation in self-regressive tree construction, it does not fully eliminate the impact of early prediction errors. Future work could explore more adaptive strategies, such as confidence-aware context selection or explicit error-correction mechanisms that allow the model to revise earlier structural decisions when later evidence contradicts them. Enabling such recovery from early mistakes may further improve robustness in long-document settings.

\bibliography{custom}

\appendix

\section{Appendix}
\subsection{Rightmost Branch Sufficiency}
We follow the setup in Figure~\ref{semantic}. Suppose that at step $i-1$, we have a partially constructed semantic tree $T_{i-1}$ and are about to process a new chunk $C_i$. If a page object $P_{child} \in C_i$ has a parent $P_{parent}$ already present in $T_{i-1}$, then $P_{parent}$ must lie on the rightmost branch of $T_{i-1}$. We show this by contradiction. If $P_{parent}$ were not on the rightmost branch, then there would exist another page object $P_{right}$ on the rightmost branch of $T_{i-1}$. Since $P_{child}$ is a child of $P_{parent}$, a depth-first traversal of $T_i$ would place $P_{child}$ before $P_{right}$. This contradicts the fact that $P_{right}$ belongs to an earlier chunk and has already been processed before chunk $C_i$. Therefore, if $P_{child}$ has a parent in $T_{i-1}$, that parent must be located on the rightmost branch of the tree.
\label{sec:appendix_rightmost}

\subsection{Error Accumulation in Self-Regressive Tree Construction}
In the ideal setting, where the semantic tree is correctly constructed, the rightmost branch suffices to capture the relevant hierarchical context. However, under realistic conditions with prediction errors, the rightmost branch may deviate from the ground-truth structure. To quantify how early prediction errors propagate in long documents, we compare performance on the first two chunks when constructing the semantic tree using either ground-truth or predicted partial trees on DocHieNet long documents, as summarized in Table~\ref{tab:error_accumulation}. F1 measures relation detection for newly introduced objects in each chunk, while TEDS evaluates the correctness of the entire semantic tree. Using ground-truth trees, the single rightmost-branch-based strategy achieves high accuracy, demonstrating that rightmost-branch selection effectively captures hierarchical structure. When using predicted trees, performance degrades substantially relative to the ground-truth setting, though it remains reasonably strong and robust.
\label{sec:appendix_error_accumulation}

\begin{table*}[h]
\centering
\begin{tabular}{l c c c c}
\hline
\textbf{Setting} & \multicolumn{2}{c}{\textbf{Chunk 0}} & \multicolumn{2}{c}{\textbf{Chunk 1}} \\
\cline{2-5}
 & F1 & TEDS & F1 & TEDS \\
\hline
Ground Truth Tree & 85.90 & 73.65 & 85.73 & 91.77 \\
Predicted Tree    & 85.90 & 73.65 & 80.56 & 72.74 \\
\hline
\end{tabular}
\caption{Error accumulation under ground-truth vs. predicted partial trees.}
\label{tab:error_accumulation}
\end{table*}

\subsection{Training and Runtime Details}
Unless otherwise specified, DOSA is trained for 100 epochs, with the learning rate decayed by a factor of 10 after 90 epochs. For the HRDoc subsets HRDS and HRDH, shorter training schedules are sufficient: models are trained for 50 epochs with the learning rate decayed after 40 epochs. This setting achieves comparable or improved performance while substantially reducing training time, suggesting that DOSA converges faster on multi-page hierarchy datasets such as HRDoc. For DocHieNet, we use a context window size of 16 for optimal performance.

All experiments are conducted on a single NVIDIA L4 GPU (24\,GB memory). The total GPU hours required for training on each dataset are as follows: DocHieNet requires approximately 46 GPU hours; HRDS and HRDH require 25 and 51 GPU hours, respectively; arXivDocs requires approximately 1 GPU hour; and E-Periodica requires approximately 2 GPU hours. To ensure reproducibility and comparability, all reported results are obtained from a single training run with a fixed random seed.
\label{sec:appendix_training}

\subsection{Dataset Preprocessing}
All document images are resized to $800 \times 800$ to facilitate visual feature extraction by the convolutional backbone. Correspondingly, bounding box annotations are rescaled to maintain correct spatial alignment. Because DOSA operates in a chunk-based manner, each document is split into chunks containing up to 256 page objects. During training, each chunk is treated as an individual training unit. During evaluation, the semantic tree is constructed incrementally by processing chunks sequentially, and context objects for each chunk are selected based on model predictions rather than ground-truth annotations to avoid providing oracle information.

Each dataset adopts its own page object annotation scheme. For instance, HRDoc defines fine-grained classes such as \textit{First-Line} and \textit{Para-Line}, while E-Periodica includes dataset-specific categories such as \textit{document\_root} and \textit{content\_block}. We preserve the original page object categories provided by each dataset and do not enforce a unified label space. As a result, DOSA employs dataset-specific category encodings during training and evaluation.

To ensure compatibility with DOSA's relation formulation, we preprocess relation annotations from each dataset to obtain a unified set of parent--child and predecessor relations. For \textbf{DocHieNet}, we directly use the provided parent--child relations and derive predecessor relations from the reading-order annotations included in the dataset. For \textbf{HRDoc}, we retain the original line-level annotations and map the dataset's \textit{contain} relations to parent--child relations and \textit{connect} relations to predecessor relations. In addition, HRDoc defines meta relations corresponding to structural roots; these are handled by assigning the associated elements as top-level parent nodes in the constructed semantic tree. For the single-page datasets \textbf{arXivDocs} and \textbf{E-Periodica}, parent--child and predecessor relations are obtained by reversing the original \textit{parent-of} and \textit{followed-by} annotations provided with the datasets. We map these annotations to the parent and predecessor relations used in DOSA. This mapping is fully reversible and preserves the original annotation information, ensuring fair and comparable evaluation across datasets.

Finally, for arXivDocs and E-Periodica, which do not include textual content annotations, we use the page object category as a semantic surrogate input.
\label{sec:appendix_dataset}

\subsection{LLM Evaluation Details}

\begin{table*}[t]
\centering
\renewcommand{\arraystretch}{1.2}  
\begin{tabular}{p{0.12\linewidth} p{0.82\linewidth}}
\hline
\textbf{Prompt} &
Here is a list whose elements represent the content blocks of a document, \newline
and the indication of keys are as follows: \\
& \textbf{"id"}: An integer that uniquely identifies the content block. \\
& \textbf{"content"}: A string representing the text in the content block. \\
& \textbf{"category"}: A string indicating the category of the content block (e.g., title, section, figure, table, etc.). \\
& \textbf{"order"}: An integer indicating the order of the content block in this document. \\
& \textbf{"page"}: An integer indicating the page number on which the content block appears. \\
& \textbf{"bbox"}: The layout information of the content block. \\
& Documents are organized as a tree-like structure. Please find the parent element\newline
of each content block based on their text and layout. \\
& The format of your reply: $[\{id_1 : parent\_id_1\}, \ldots, \{id_n : parent\_id_n\}]$. \newline
Do not include any other content. \\
& Here are some demonstration: \{Demonstrates\} \\
& Here is the input document: \{Elements\} and \{Images\} \\
& --- \\
& reply: \\
\hline
\textbf{Slots} & \textbf{Elements}: List of document layout entities from DocHieNet. \\
& \textbf{Images}: List of document images in base64 format. \\
& \textbf{Demonstrates}: The selected demonstration with ground truth response. \\
\hline
\end{tabular}
\caption{The prompt for evaluating LLMs on DocHieNet.}
\label{tab:prompt_dochienet}
\end{table*}

We adopt the prompt formulation introduced in DocHieNet and extend it along two dimensions to align with our model's evaluation setup: (1) multimodal input — in addition to page object annotations such as ordering, bounding boxes, and content, we provide document images to the multimodal LLM, enabling the model to leverage visual layout cues; and (2) structured output enforcement — the model is required to generate parent-child relations in a predefined Pydantic schema. A representative prompt is shown in Table~\ref{tab:prompt_dochienet}.

\label{sec:appendix_llm}

\begin{table}[h]
\renewcommand{\arraystretch}{1.2}
\centering
\begin{tabular}{c|cc}
\hline
\textbf{Window Size} & \textbf{F1} & \textbf{TEDS} \\
\hline
0  & 81.61 & 56.30 \\
8  & 83.56 & 72.23 \\
16 & 83.97 & \textbf{73.68} \\
24 & \textbf{84.05} & 73.56 \\
32 & 83.99 & 72.97 \\
\hline
\end{tabular}
\caption{Context window size sensitivity on DocHieNet.}
\label{tab:context_window}
\end{table}

\subsection{Window Size Sensitivity}
We further analyze the impact of the window size for the rightmost-branches-based soft window mechanism by varying the number of context objects retrieved from the partial semantic tree when processing a new chunk. As shown in Table~\ref{tab:context_window}, performance improves as the window size increases, with both F1 and TEDS stabilizing around a window size of 16 on long documents in DocHieNet. This suggests that most structurally relevant long-range dependencies are concentrated within recent branches of the tree frontier.

We note that the optimal window size is influenced by dataset characteristics, including document length, structural depth, and cross-chunk dependency patterns. In datasets with deeper hierarchies or denser inter-chunk relations, a larger window may provide additional robustness, while excessively large windows may introduce irrelevant context and slight performance degradation. These observations indicate that the window size reflects a data-dependent trade-off between contextual coverage and noise.
\label{sec:appendix_window_size}

\subsection{Sizing Embedding Analysis}
\begin{figure*}[t]
  \centering
   \includegraphics[width=\textwidth]{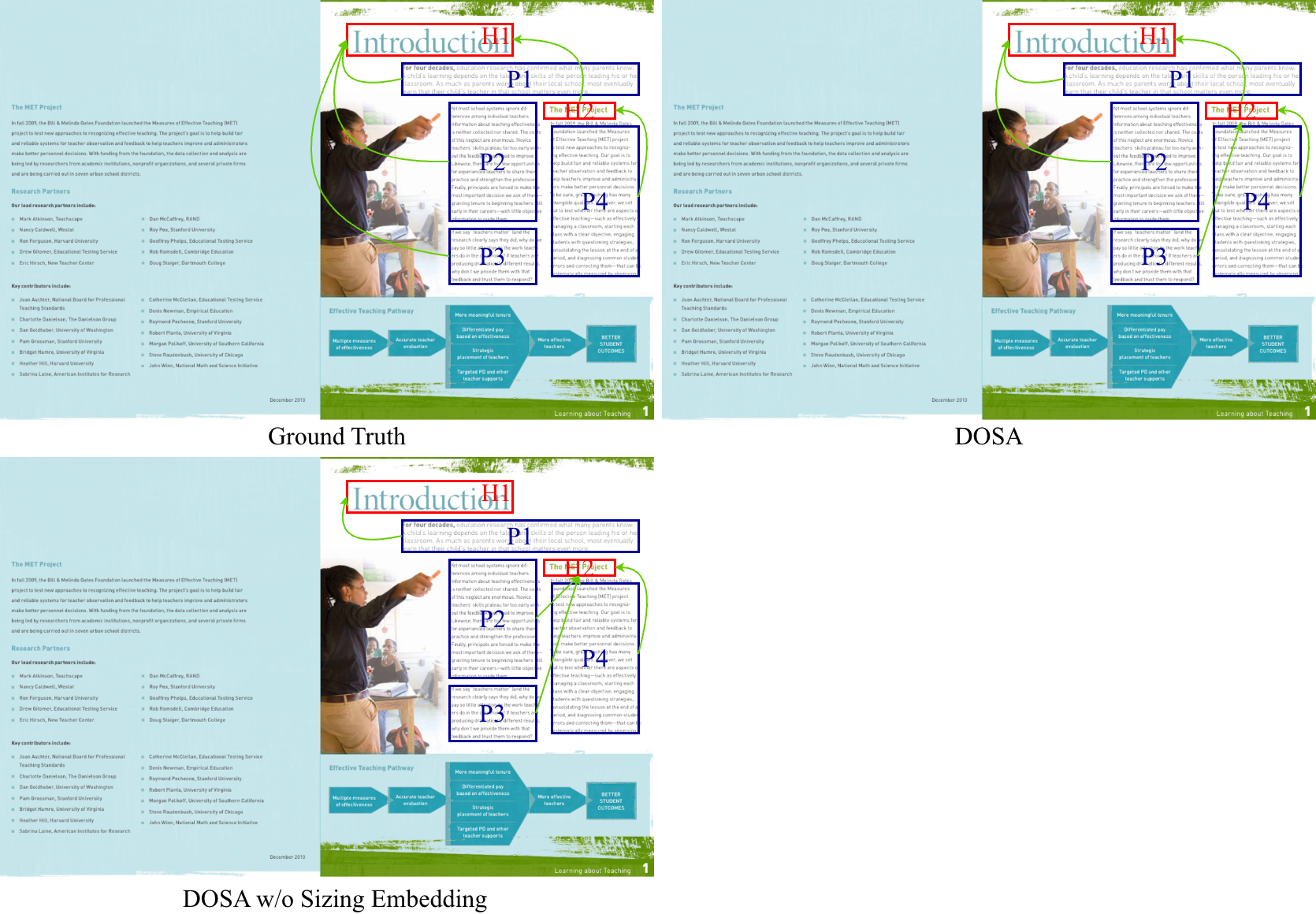}
  \caption{Qualitative comparison of size embedding in complex layout.}
  \label{fig:qualitative}
\end{figure*}

Size cues can be particularly informative for distinguishing between different levels of section headers, especially when explicit section numbers are absent. However, in many cases, hierarchical relations can be inferred from multiple complementary signals—such as section numbering patterns, structural layout, and positional information—thereby reducing reliance on font-size cues alone. To further examine the contribution of size embeddings, we conducted a qualitative analysis on documents with complex layouts, where size information may play a more critical role in accurate hierarchy prediction. As shown in Figure~\ref{fig:qualitative}, the example document has an unusually wide format with multiple object columns that are not strictly aligned column-by-column, resulting in a complex layout structure. Overall, the model incorporating size embeddings achieves higher detection accuracy than the model without them. Notably, the model without size embeddings misclassifies H2 as a top-level section header, whereas the model with size embeddings correctly identifies it as a child of H1.

\label{sec:appendix_sizing_analysis}

\subsection{Inference Time Analysis}
DOSA processes documents in a self-regressive manner, where dependencies between chunks prevent full parallelization within a single document. This sequential dependency could potentially increase per-document inference latency. To provide empirical context, we evaluated DOSA on the DocHieNet benchmark, which contains 161 documents. Across a total of 208 chunks (11 pages per chunk on average), the full runtime was 206.18 seconds on a single NVIDIA L4 GPU, without parallelization across chunks. This corresponds to an average of 0.99 seconds per chunk and approximately 1.28 seconds per document. Although self-regressive processing introduces sequential overhead, the latency remains moderate in practice. Furthermore, in production settings, the overhead can be amortized by processing multiple documents in parallel. Therefore, DOSA remains practical for real-world document processing scenarios.
\label{sec:appendix_inference_time}

\end{document}